\documentclass[lettersize,journal]{IEEEtran}
\usepackage{amsmath,amsfonts}
\usepackage{algorithmic}
\usepackage{algorithm}
\usepackage{array}
\usepackage[caption=false,font=normalsize,labelfont=sf,textfont=sf]{subfig}
\usepackage{textcomp}
\usepackage{stfloats}
\usepackage{url}
\usepackage{verbatim}
\usepackage{graphicx}
\usepackage{cite}
\usepackage[colorlinks=true,allcolors=blue]{hyperref}
\usepackage{booktabs} % 需在导言区引入 booktabs 宏包
\makeatletter
\usepackage{adjustbox} % 导言区引入
\makeatother
\usepackage{graphicx}
\usepackage[table]{xcolor}
\usepackage{tabu}                     % 表格插入
\usepackage{multirow}                 % 一般用以设计表格，将所行合并
\usepackage{multicol}                 % 合并多列
\usepackage{float}                    % 图片浮动
\usepackage{makecell}                 % 三线表-竖线
\usepackage{booktabs}                 % 三线表-短细横线
\usepackage{cleveref}
\makeatletter
\def\@cite#1#2{{\color{blue}[\textbf{#1\if@tempswa , #2\fi}]}} % 将 citecolor 改为 blue
\makeatother

\crefname{figure}{fig}{figures}
\Crefname{figure}{Fig}{Figures}
\Crefname{equation}{Eq}{Equation}
\crefname{table}{tab}{tables}      % 小写引用格式
\Crefname{table}{Tab}{Tables}      % 句首引用格式

\begin{document}

\title{MSD-KMamba: Bidirectional Spatial-Aware Multi-Modal 3D Brain Segmentation via Multi-scale Self-Distilled Fusion Strategy}

\author{Dayu Tan,
Ziwei Zhang,
Yansan Su,
Xin Peng,
Yike Dai,
Chunhou Zheng,
and Weimin Zhong
        % <-this % stops a space
\thanks{This work was supported in part by the National Key Research and Development Program of China (2021YFE0102100), in part by National Natural Science Foundation of China (62303014, 62172002, 62322301). (\emph{Corresponding author: Yansen Su.})}
\thanks{Dayu Tan, Ziwei Zhang, Yansen Su, and Chunhou Zheng are with the Key Laboratory of Intelligent Computing and Signal Processing, Ministry of Education, Anhui University, Hefei 230601, China (e-mail: suyansen@ahu.edu.cn).}
\thanks{Xin Peng and Weimin Zhong are with the Key Laboratory of Smart Manufacturing in Energy Chemical Process, Ministry of Education, East China University of Science and Technology, Shanghai 200237, China (e-mail: wmzhong@ecust.edu.cn).}
\thanks{Yike Dai is with the Department of Orthopedic, Beijing Friendship Hospital, Capital Medical University, Beijing 100050, China (e-mail: dai\_yike@126.com)}
}


\maketitle

\begin{abstract}
Numerous CNN-Transformer hybrid models rely on high-complexity global attention mechanisms to capture long-range dependencies, which introduces non-linear computational complexity and leads to significant resource consumption. Although knowledge distillation and sparse attention mechanisms can improve efficiency, they often fall short of delivering the high segmentation accuracy necessary for complex tasks. Balancing model performance with computational efficiency remains a critical challenge. In this work, we propose a novel 3D multi-modal image segmentation framework, termed MSD-KMamba, which integrates bidirectional spatial perception with multi-scale self-distillation fusion strategy. The bidirectional spatial aware branch effectively captures long-range spatial context dependencies across brain regions, while also incorporating a powerful nonlinear feature extraction mechanism that further enhances the model’s ability to learn complex and heterogeneous patterns. In addition, the proposed multi-scale self-distilled fusion strategy strengthens hierarchical feature representations and improves the transfer of semantic information at different resolution levels. By jointly leveraging the bidirectional spatial perception branch and the multi-scale self-distilled fusion strategy, our framework effectively mitigates the bottleneck of quadratic computational complexity in volumetric segmentation, while simultaneously addressing the limitation of insufficient global perception. Extensive experiments on multiple standard benchmark datasets demonstrate that MSD-KMamba consistently outperforms state-of-the-art methods in segmentation accuracy, robustness, and generalization, while maintaining high computational efficiency and favorable scalability. The source code of MSD-KMamba is publicly available at https://github.com/daimao-zhang/MSD-KMamba.
\end{abstract}


\section{Introduction}
\IEEEPARstart{B}{rain} medical image segmentation in 3D plays a vital role in modern clinical workflows by enabling the precise extraction of anatomical and pathological structures from volumetric modalities such as computed tomography (CT) and magnetic resonance imaging (MRI) \cite{ilesanmi2024reviewing}. The segmentation process is foundational for various applications, including tumor delineation, organ localization, surgical navigation, and radiotherapy planning. In recent years, deep learning-based segmentation methods have become the dominant paradigm. Convolutional neural networks (CNNs)-based architectures such as 3D U-Net \cite{cciccek20163d} and V-Net \cite{milletari2016v}, specifically designed for medical volumetric image segmentation, have demonstrated significant success through encoder-decoder structures with skip connections, effectively capturing multi-scale spatial features within volumetric data.

% Despite their effectiveness, CNNs are inherently limited in modeling long-range dependencies due to their local receptive \cite{linsley2018learning}. 
Despite the strong performance of CNNs in segmenting local spatial features, they are constrained by their inability to capture global context \cite{linsley2018learning}.
Transformer-based models have been introduced to address the limited receptive field of CNNs by leveraging global self-attention mechanisms \cite{xiao2023transformers}, allowing for improved contextual understanding across spatial dimensions. Nevertheless, the quadratic computational complexity of standard Transformers becomes particularly problematic in 3D scenarios, where volumetric inputs impose a heavy burden on memory and processing. To overcome the high computational cost of self-attention, Mamba—a recently proposed selective state-space model—offers an efficient alternative. By employing parameterized state dynamics, Mamba captures long-range dependencies with linear time and memory complexity \cite{gu2023mamba}. These efficiency gains make the model highly attractive for 3D medical segmentation tasks that require both global reasoning and computational scalability. Concurrently, Kolmogorov–Arnold networks \cite{liu2024kan} enable efficient segmentation through universal function approximation, improving robustness to low contrast and partial volume effects without deep architectures.

While the linear-complexity selective state-space mechanism in Mamba improves efficiency, deep learning models still face overfitting and unstable convergence on complex medical data. Self-distillation has emerged as a lightweight strategy to enhance robustness and generalization, eliminating the need for external teachers \cite{zhang2021self}. It enables a network to refine its own predictions by transferring knowledge from deeper to shallower layers, promoting consistent representations and stabilizing training. Recent approaches like dual self-distillation \cite{zhang2025dualdistill} further improve this by incorporating guidance from both encoder and decoder pathways, enabling multi-level feature refinement without adding complexity. Combined with Mamba, self-distillation offers a powerful framework for building high-performance, generalizable 3D segmentation models suited for diverse clinical settings.

In this work, we present MSD-KMamba, a novel network architecture designed to bridge semantic gaps across imaging modalities and improve segmentation robustness via a Multi-scale Self-distillation Aggregation (MDA) mechanism. To achieve efficient long-range dependency modeling, a Bidirectional Kan-enhanced Mamba (BKM) module is integrated, operating with linear computational complexity. Nevertheless, multi-modal feature integration continues to pose challenges such as semantic misalignment and restricted receptive fields. To tackle these issues, a Hierarchical Semantic Alignment (HSA) block is incorporated, enabling more coherent and context-aware fusion of cross-modal information. Collectively, these components enhance the model’s representational capacity while substantially reducing computational overhead. The main contributions of this work are summarized as follows:
\begin{itemize}
\item{We design the MDA mechanism that employs a self-distillation framework. Within this framework, a deep teacher network captures rich semantic representations through a multi-scale fusion attention mechanism, while a shallow student network inherits these high-level semantics via soft labels and feature distillation. }
\item{We propose a novel BKM method integrating bidirectional global modeling with a principled nonlinear transformation. It captures contextual dependencies via simultaneous forward and backward scanning of volumetric data, significantly enhancing representation of complex structural and cross-dimensional interactions with high efficiency.}
\item{This study presents the HSA module to resolve semantic and receptive field limitations in multi-modal fusion. It combines atrous large-kernel convolution for broad context with separable convolution for efficiency. The innovation lies in a cross-modality attention gate and multi-level feature calibration, enabling semantics-aware integration with low computational cost.
}
\end{itemize}

The structure of this study is as follows: In Section \ref{sec:related}, related works about 3D medical image segmentation, Mamba module and self-distillation are introduced. The methodology of our proposed model is detailed in Section \ref{sec:proposed}. Elaborate experiments and analysis are conducted in Section \ref{sec:experiments}. Finally, Section \ref{sec:conclusion} presents the summarization of our research conclusions.

\section{RELATED WORK}\label{sec:related}
\subsection{3D Medical Image Segmentation}
3D medical image segmentation plays a critical role in medical image analysis, focusing on the precise delineation of anatomical structures or pathological regions—such as organs, tumors, and tissues—across volumetric data. Conventional segmentation methods, including thresholding, region growing, and deformable models, rely extensively on hand-crafted features and strong domain-specific assumptions \cite{xu2024advances}. While these techniques can perform well in constrained settings, they often fail to generalize across diverse imaging modalities, variable image quality, and complex anatomical presentations encountered in real-world clinical practice.

The advent of deep learning has revolutionized the field of medical image segmentation, with CNNs becoming the backbone of most modern segmentation models. Architectures such as U-Net and its 3D extension (3D U-Net) have shown excellent performance by leveraging encoder–decoder structures and skip connections to preserve spatial resolution \cite{tan2024novel}. Subsequent variants like V-Net, Attention U-Net \cite{khan2020survey}, and nnU-Net \cite{isensee2021nnu} introduced improvements such as volumetric convolutions, attention mechanisms, and automated network configuration, respectively. However, CNNs are inherently limited in modeling long-range dependencies due to their localized receptive fields, which can be insufficient for understanding complex global anatomical contexts.

To address the limitation of modeling long-range dependencies, Transformer-based models have been introduced into 3D medical image segmentation. Architectures like TransBTS \cite{wang2021transbts}, nnFormer \cite{zhou2023nnformer}, and UNETR \cite{hatamizadeh2022unetr} incorporate self-attention mechanisms to capture long-range dependencies and global semantic context, leading to improved segmentation accuracy in complex scenarios. Nevertheless, the high computational and memory demands of pure Transformer architectures have restricted their scalability and deployment in clinical practice. Recent models aim to strike a balance between accuracy and efficiency. For instance, hybrid and lightweight designs such as VT-UNet \cite{peiris2022robust} and SegFormer3D \cite{perera2024segformer3d} combine CNN backbones with efficient Transformer modules to reduce computational burden without compromising segmentation quality. Despite these advances, challenges remain in handling low-contrast images, domain shifts across modalities, and limited annotation. Future research will likely focus on efficient architectures, robust multi-modal fusion, and semi-/self-supervised learning \cite{tan2023deep} strategies to enhance both generalizability and clinical applicability.

\subsection{Mamba’s Selective State Space Model}
The initial application of Mamba’s State Space Model (SSM) in the field of medical image segmentation aimed to overcome the inherent limitations of traditional CNNs and Transformers. Early studies revealed that CNNs often struggle to capture long-range dependencies between anatomical regions, while the high computational complexity of Transformers impedes their use in high-resolution volumetric data \cite{khan2022transformers}. The LKM-UNet \cite{wang2024lkm}, which integrates large kernel convolutions and Mamba modules into a U-Net-like framework, has demonstrated the potential benefits of combining extended receptive fields with efficient sequence modeling in medical image segmentation tasks. Subsequent developments, such as ShapeMamba-EM \cite{shi2024shapemamba}, have progressively enhanced the representation of both local and global features by combining local shape priors and dynamic state space modeling, especially in fine-grained electron microscopy (EM) segmentation tasks.

EM-Net \cite{chang2024net} builds upon earlier models by incorporating efficient channel and frequency domain processing into Mamba blocks, significantly improving memory efficiency and speed while preserving rich multi-scale anatomical context. Meanwhile, SegMamba \cite{xing2024segmamba} introduces a novel 3D graph construction mechanism that simultaneously models spatial dependencies across axial, coronal, and sagittal planes, thereby boosting the detection accuracy of small and complex structures such as the pancreatic duct. In parallel, PathMamba \cite{fan2024pathmamba} leverages weak supervision and multi-scale state space modeling for effective multi-class segmentation in histopathology images, demonstrating that Mamba-based approaches can generalize well to different imaging modalities and supervision levels.

The hierarchical gated convolution and Mamba-based U-Net architecture of MambaClinix \cite{bian2024mambaclinix} exemplifies SSMs' ability to model long-range dependencies with linear computational complexity, positioning SSMs as a powerful and efficient alternative to traditional attention-based methods in medical image segmentation. Overall, Mamba-based SSM architectures have evolved from 2D segmentation tasks to fully volumetric 3D anatomical modeling and have expanded from single-modality inputs to multi-modal and weakly-supervised fusion strategies.

\subsection{Self-distillation in Medical Image Segmentation}
Self-distillation has emerged as a powerful technique for boosting both segmentation accuracy and training efficiency in medical image analysis. One of the earliest frameworks, Dual Self-Distillation (DSD) \cite{banerjee2024dual}, introduces bi-directional knowledge transfer between the encoder and decoder of a U-shaped network. Through multi-scale supervision, DSD significantly enhances segmentation performance on challenging anatomical structures such as small or low-contrast organs, while maintaining low computational cost during inference.

To improve spatial and relational consistency, MSKD \cite{zhao2023mskd} introduces multi-scale structured knowledge distillation via feature filtering and graph-based region alignment. The approach enables more robust representation learning between teacher and student networks, improving generalization under complex anatomical variations. MISSU \cite{wang2023missu} extends self-distillation to hybrid CNN-Transformer models by incorporating a self-distilled decoder and skip fusion. The method achieves competitive accuracy on volumetric benchmarks like BraTS and CHAOS with minimal training overhead, striking a strong balance between expressiveness and efficiency.

More recently, Efficient Knowledge Distillation for Brain Tumor Segmentation \cite{qi2022efficient} focuses on boundary-aware objectives to enforce consistency in tumor margins and interiors—crucial for clinical relevance. Meanwhile, SMIT \cite{jiang2022self} introduces self-distillation into masked image Transformer pretraining, enabling dense anatomical prior learning from unlabeled volumes. The self-distillation methodology significantly boosts downstream segmentation accuracy without affecting inference latency, making it ideal for scalable clinical deployment. Collectively, self-distillation advancements illustrate how self-distillation has evolved into a lightweight yet highly effective strategy for building accurate, efficient, and generalizable medical segmentation models.

\section{PROPOSED METHOD}\label{sec:proposed}
In this section, the proposed MSD-KMamba includes an encoder, a skip connection layer, and a decoder. First, we describe the overall process of MSD-KMamba, and then elaborate on each component of MSD-KMamba.
\subsection{MSD-KMamba for Medical Image Segmentation}
Fig. \ref{fig:model1-overview} illustrates the overall framework of MSD-Kmamba. During the downsampling process, the Hierarchical Semantic Alignment (HSA) module is employed. Large-kernel depthwise convolutions provide strong long-range modeling capability, while channel-separated convolutions offer feature decoupling and enhanced representation. The combination of both significantly improves model performance and generalization under efficient computation, making it especially suitable for medical image segmentation tasks involving complex structures and multi-modal features. When input resolution is low, the Bidirectional Kan-enhanced Mamba (BKM) module is applied to capture spatial context and long-range dependencies. The integration of kernel functions further enhances the semantic richness of attention, improving the modeling of complex feature patterns.

In the middle part of the constructed network, a novel Multi-scale Self-distillation Aggregation (MDA) module is designed. The fused multi-scale features contain information at different granularities. On the basis of multi-scale feature fusion, a self-distillation mechanism performs knowledge compression and transfer, which enhances the discriminative ability of critical regions (e.g., boundaries and lesions) while maintaining global structural understanding. The distilled multi-scale hierarchical features are then progressively passed to the decoder, improving segmentation accuracy for small targets and ambiguous regions.

\subsection{MDA: Multi-scale Self-distillation Aggregation}
\begin{figure*}[!t]
\centering
\includegraphics[width=\textwidth,height=0.49\textheight]{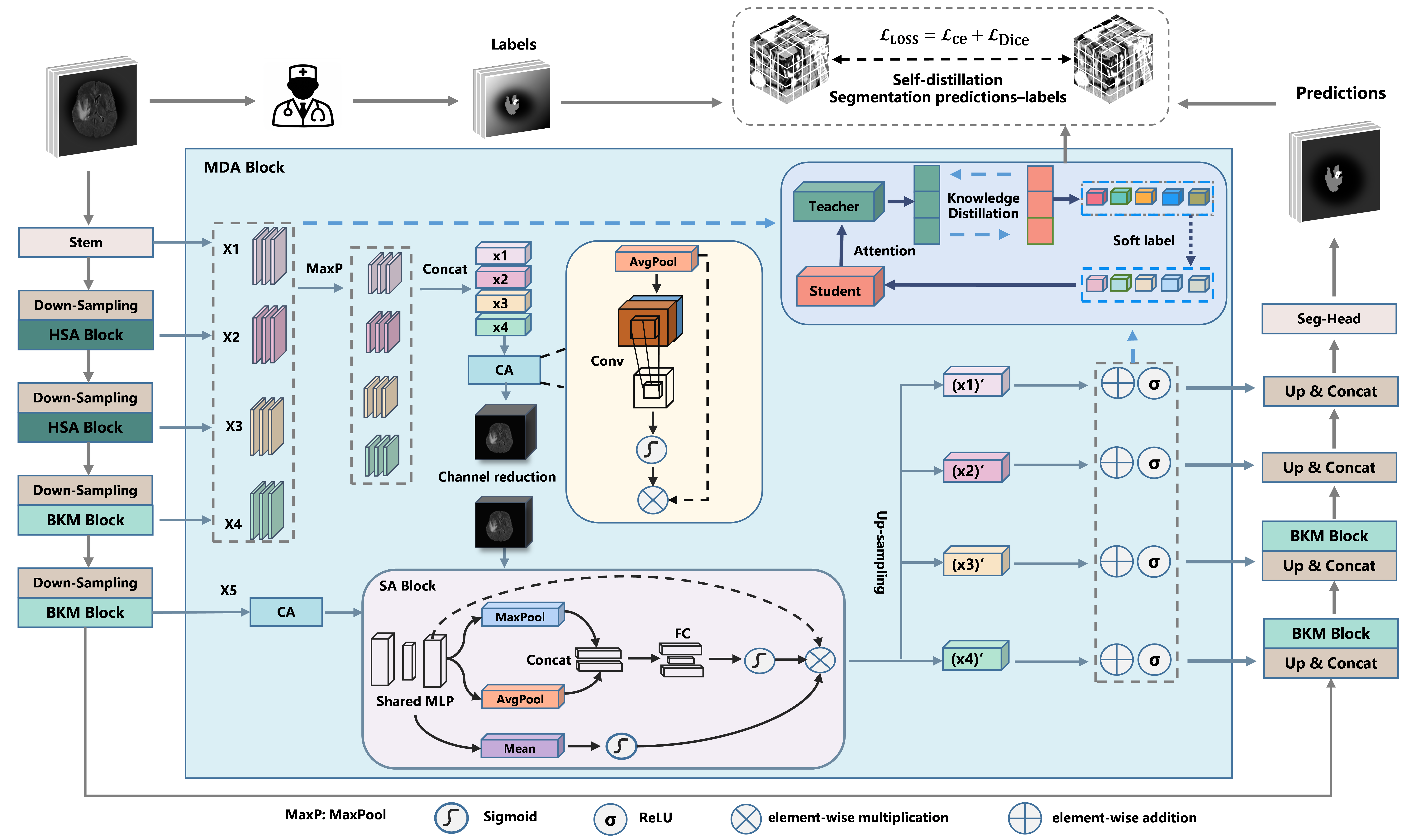}
\caption{Overview of the Network Architecture. The MDA module fuses multi-scale contextual features from the encoder using attention mechanisms and refines them via self-distillation to reduce computational cost. The BKM module models long-range dependencies and nonlinear feature transformations to fully capture feature relationships. The HSA module integrates features from different modalities using large-kernel depthwise convolutions and channel-separated convolution strategies.}
\label{fig:model1-overview}
\end{figure*}

We design a novel MDA module to address the semantic inconsistency commonly observed in multi-scale feature fusion, as shown in Fig. \ref{fig:model1-overview}. While conventional fusion strategies integrate low-level spatial details with high-level semantic features, they often suffer from feature misalignment and semantic ambiguity across scales. By introducing a self-distillation mechanism, MDA enables the network to extract high-quality, self-supervised semantic knowledge from the aggregated features, guiding shallow and deep representations toward mutual consistency and enhancing the overall semantic coherence of the network. The fusion process is formally defined as:

\begin{equation}
\eta = \mathcal{F}_{concat} \left( \bigoplus_{k=1}^{4} \mathcal{D}^{(k)}({X}^{(k)}) \right),
\end{equation}
\noindent
in Eq. (1), $\mathcal{F}_{concat}(\cdot) $ denotes the feature fusion operator implemented through channel-wise concatenation, $\mathcal{D}^{(k)}: \mathbb{R}^{C_k \times H_k \times W_k \times D_k} \to \mathbb{R}^{C_k \times H_5 \times W_5 \times D_5}$ represents a parametric downsampling transformation that progressively reduces the spatial dimensions of the $k$-th scale feature to match the deepest level ${X}^{(5)}$, and ${X}^{(k)} \in \mathbb{R}^{C_k \times H_k \times W_k \times D_k}$ constitutes the hierarchical multi-scale contextual features extracted from the encoder pathway, with spatial dimensions satisfying $H_k > H_{k+1}$, $W_k > W_{k+1}$, $D_k > D_{k+1}$ for $k = 1,2,3,4$.

\begin{equation}
\nu = \mathcal{A}{\text{spatial}} \left( \mathcal{P}(\mathcal{A}{c}(\eta)) \right) \oplus \mathcal{A}{\text{spatial}} \left( \mathcal{A}{c}({X}^{(5)}) \right),
\end{equation}
\noindent
where $\mathcal{A}{c}(\cdot)$ and $\mathcal{A}{\text{spatial}}(\cdot)$ represent channel-wise and spatial attention mappings respectively, $\mathcal{P}(\cdot)$ indicates a channel projection operation that reduces the channel dimension to match ${X}^{(5)}$, and $\oplus$ denotes element-wise summation. Specifically, $\mathcal{A}{c}(\cdot): \mathbb{R}^{C \times H \times W \times D} \to \mathbb{R}^{C \times H \times W \times D}$, $\mathcal{A}{\text{spatial}}(\cdot): \mathbb{R}^{C \times H \times W \times D} \to \mathbb{R}^{C \times H \times W \times D}$, and $\mathcal{P}(\cdot): \mathbb{R}^{C \times H \times W \times D} \to \mathbb{R}^{C_5 \times H \times W \times D}$.

\begin{equation}
{X}^{\text{out}}_i = \sigma \left( \mathcal{U}_i({P}_i(\nu))\oplus {X}^{(i)} \right), \quad \forall i \in {1,2,3,4},
\end{equation}
\noindent
in Eq. (3), $\mathcal{U}_i(\cdot): \mathbb{R}^{C \times H \times W \times D} \to \mathbb{R}^{C \times H_i \times W_i \times D_i}$ denotes a scale-specific spatial upsampling operator that restores the spatial resolution of $\nu$ to match that of ${X}^{(i)}$, and $\mathcal{P}_i(\cdot): \mathbb{R}^{C \times H_i \times W_i \times D_i} \to \mathbb{R}^{C_i \times H_i \times W_i \times D_i}$ represents a subsequent channel projection that adjusts the channel dimension to match ${X}^{(i)}$. $\sigma(\cdot)$ denotes the ReLU activation function. The refined feature tensor ${X}^{\text{out}}_i \in \mathbb{R}^{C_i \times H_i \times W_i \times D_i}$ is subsequently propagated to the decoder and further engaged in a self-distillation process with the original encoder feature ${X}^{(i)}$ to facilitate semantic knowledge transfer across scales.

 The Softmax operation is computed over the channel dimension at each voxel location. The fused multi-scale features are utilized to generate soft labels or auxiliary supervision, which further enhances segmentation performance while maintaining the advantage of lightweight deployment. The probabilistic outputs for both the original and refined features are obtained through the following Softmax transformations:

\begin{equation}
\hat{x}^{\text{out}}_{i, \kappa} = \frac{e^{x^{\text{out}}{i, \kappa}}}{\sum\limits_{\xi=1}^{\mathcal{C}} e^{x^{\text{out}}{i, \xi}}}, \quad
\hat{x}_{i, \kappa} = \frac{e^{x_{i, \kappa}}}{\sum\limits_{\xi=1}^{\mathcal{C}} e^{x_{i, \xi}}},\quad (i = 1, 2, 3, 4),
\end{equation}

\noindent
where $\kappa$ and $\xi$ index over the class channels, and $\mathcal{C}$ denotes the total number of semantic classes.

Compared with conventional knowledge distillation methods that require an additional complex teacher network, self-distillation is performed entirely within the model, aligning and optimizing features without introducing extra computational cost during inference. The distributional matching objective between teacher and student pathways is defined as:

\begin{equation}
\begin{aligned}
\mathcal{L}_{Distribution}^{(i)} &= - \sum_{\theta=1}^{C} \hat{x}_{i, \theta} \log(\hat{x}_{i, \theta}^{\text{out}}),
\end{aligned}
\end{equation}
in Eq. (5), minimizing the informational divergence between probability distributions of the teacher and student pathways ensures effective transfer of probabilistic knowledge across hierarchical feature levels while preserving gradient stability during optimization. To enhance geometric consistency between representations, we introduce a structural preservation term:

\begin{equation}
\begin{aligned}
\mathcal{L}_{Struct}^{(i)} &= 1 - \frac{2 \sum_{\theta=1}^{C} \hat{x}_{i, \theta} \hat{x}_{i, \theta}^{\text{out}}}{\sum_{\theta=1}^{C} \hat{x}_{i, \theta} + \sum_{\theta=1}^{C} \hat{x}_{i, \theta}^{\text{out}}}.
\end{aligned}
\end{equation}

Herein, quantifying the feature-level structural consensus between original and refined representations enhances geometric consistency and region-aware alignment across multi-scale feature maps. In the context of 3D medical image segmentation, such a structural preservation term effectively maintains organ morphology and lesion topology throughout the distillation process. The complete self-distillation objective combines both components through a balanced weighting scheme:

\begin{equation}
\begin{aligned}
\mathcal{L}_{SD} &= \sum_{i=1}^{4} \left\{ \alpha \cdot \left[ 1 - \frac{2 \sum_{\theta=1}^{C} \hat{x}_{i, \theta} \hat{x}_{i, \theta}^{\text{out}}}{\sum_{\theta=1}^{C} \hat{x}_{i, \theta} + \sum_{\theta=1}^{C} \hat{x}_{i, \theta}^{\text{out}}} \right] \right. \\
&\quad \left. + (1 - \alpha) \cdot \left[ -\sum_{\theta=1}^{C} \hat{x}_{i, \theta} \log(\hat{x}_{i, \theta}^{\text{out}}) \right] \right\},
\end{aligned}
\end{equation}
in Eq. (7), the unified self-distillation loss \(\mathcal{L}_{SD}\) constitutes a convex combination of structural consensus and distributional matching components, parameterized by \(\alpha \in [0,1]\), which enables simultaneous optimization of both probabilistic similarity and geometric fidelity across all feature scales without introducing additional computational overhead during inference.

\begin{figure*}[t!]
\centering
\includegraphics[width=1\textwidth,height=5.35cm]{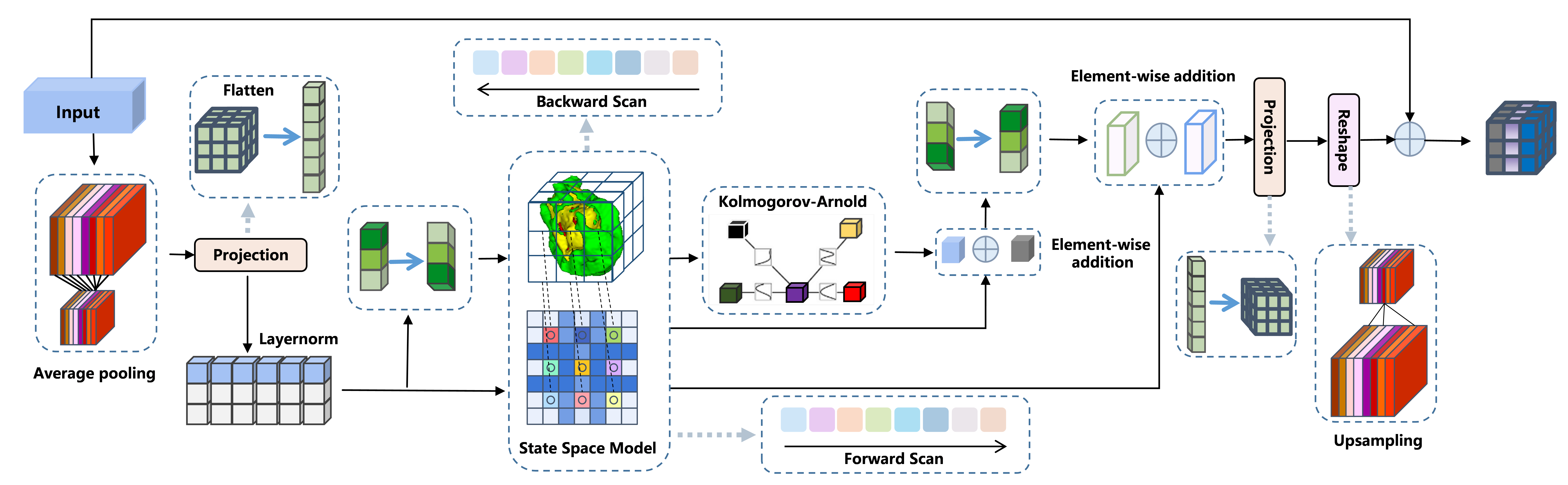}
\caption{Bidirectional Kan-enhanced Mamba (BKM) module mainly consists of a forward Mamba branch and a backward branch integrated with the additive structure of nonlinear function transformations.}
\label{fig:BKM}
\end{figure*}

\subsection{BKM: Bidirectional Kan-enhanced Mamba}

We propose a BKM block, as illustrated in Fig. \ref{fig:BKM}. BKM module integrates complementary spatial aware for dual-directional feature modeling. The forward branch captures long-range contextual dependencies and enhances semantic representations, while the backward branch propagates high-level semantics to lower layers for detail preservation and boundary refinement. An adaptive spectral network further enhances nonlinear modeling capacity through learnable basis functions, enabling expressive feature transformation. The bidirectional spatial aware process is formally defined as:

\begin{equation}
{F}_f = \mathcal{T}_f({F}_0; \Theta_f),  \
{F}_b = \mathcal{T}_b({F}_0; \Theta_b),
\end{equation}

% \begin{cases}
% \mathcal{F}_f = \mathcal{T}_f(\mathcal{F}_0; \Theta_f), \\
% \mathcal{F}_d = \mathcal{T}_d(\mathcal{F}_0; \Theta_d)
% \end{cases}

\noindent
where ${F}_0 \in \mathbb{R}^{C \times H \times W \times D}$ is the projected intermediate feature. $\mathcal{T}_f$ and $\mathcal{T}_b$ denote complementary transformation networks parameterized by $\Theta_f$ and $\Theta_b$, designed to capture long-range contextual dependencies and enable multi-scale feature propagation with structural refinement, respectively. The feature fusion mechanism combines the transformed representations through:

\begin{equation}
{F}_{fuse} = {F}_f + \varsigma\left(\mathcal{N}({F}_b); \Omega\right) + {F}_b.
\end{equation}
\noindent

Herein, $\mathcal{N}(\cdot)$ indicates layer normalization, while $\varsigma(\cdot; \Omega): \mathbb{R}^{C \times H \times W \times D} \to \mathbb{R}^{C \times H \times W \times D}$ denotes a parametric nonlinear operator with learnable basis functions parameterized by $\Omega$, enhancing representational capacity through adaptive spectral decomposition. Such fusion mechanism facilitates simultaneous integration of global semantic context and local structural details. 
The final output is obtained through a residual connection:

\begin{equation}
{Y} = \mathcal{P}'({F}_{fuse}; \Theta_p) + {X} ,
\end{equation}
\noindent
where the projection mapping $\mathcal{P}'(\cdot; \Theta_p): \mathbb{R}^{C \times (H \times W \times D)} \to \mathbb{R}^{C \times H \times W \times D}$, parameterized by $\Theta_p$, restores the original feature dimensionality. The output ${Y}$ maintains identical spatial resolution as the input ${X}$, ensuring precise spatial correspondence required for dense prediction tasks.

The proposed bidirectional architecture enables synergistic interaction between global context modeling and detailed feature reconstruction through dual complementary transformation pathways. The incorporation of an adaptive nonlinear operator with learnable basis functions significantly enhances the model's capacity to capture complex anatomical structures and pathological variations in medical imaging data. 
The spectral decomposition is mathematically represented as:

\begin{equation}
y = \sum_{i=1}^{N} \mu_i \cdot \mathcal{K}_i({x}; {\theta}_i),
\end{equation}
\noindent
in Eq. (11), $\mathcal{K}_i$ denotes a family of parameterized univariate kernel functions (e.g., adaptive spline basis or polynomial kernels) with learnable parameters ${\theta}_i$, and $\mu_i \in \mathbb{R}$ represents the corresponding spectral coefficient. Nonlinear transformations are intrinsically embedded within the kernel representations, eliminating the need for explicit activation functions while enhancing representational efficiency. The theoretical foundation of our approach is grounded in the decomposable nonlinear representation theorem:

\begin{equation}
f({x}) = \sum_{q=1}^{Q} \phi_q \left( \sum_{p=1}^{P} \psi_{q,p}(x_p) \right),
\end{equation}
\noindent
where this universal decomposition theorem establishes that any continuous multivariate function can be decomposed into a finite composition of continuous univariate functions $\phi_q$ and $\psi_{q,p}$. The proposed architecture leverages this adaptive function composition framework to construct a learnable nonlinear operator that efficiently models complex feature interactions in high-dimensional spaces, particularly beneficial for capturing intricate anatomical patterns in volumetric medical imagery.
The state-space modeling framework is formulated as:

\begin{equation}
\left\{
\begin{aligned}
{s}_t &= {\Lambda} \cdot {s}_{t-1} + {\Gamma} \cdot {u}_t,
\\
{w}_t &= {\tau} \cdot {s}_t,
\end{aligned}
\right.
\end{equation}

\noindent
where ${s}_t \in \mathbb{R}^D$ is the state vector, ${u}_t \in \mathbb{R}^D$ the input features, and ${w}_t \in \mathbb{R}^D$ the output. Learnable matrices ${\Lambda}$, ${\Gamma}$, and ${\tau}$ control state transitions, input integration, and output mapping. Such state-space formulation efficiently captures long-range spatial dependencies via iterative state propagation, benefiting segmentation of anatomies with extended contextual relationships in 3D medical images.

The integration of spectral kernels with state-space modeling offers a powerful framework for medical image segmentation. Kernel-based operators capture complex nonlinear feature transformations through learnable basis functions, while state-space formulations maintain spatiotemporal coherence in volumetric data. The hybrid approach effectively addresses heterogeneous tissue appearance, ambiguous boundaries, and multi-scale context preservation, particularly in MRI and CT where intensity variations and structural complexity demand advanced modeling capabilities.

\begin{figure}[t!]
\centering
\includegraphics[width=1\linewidth,height=0.7\textheight,keepaspectratio]{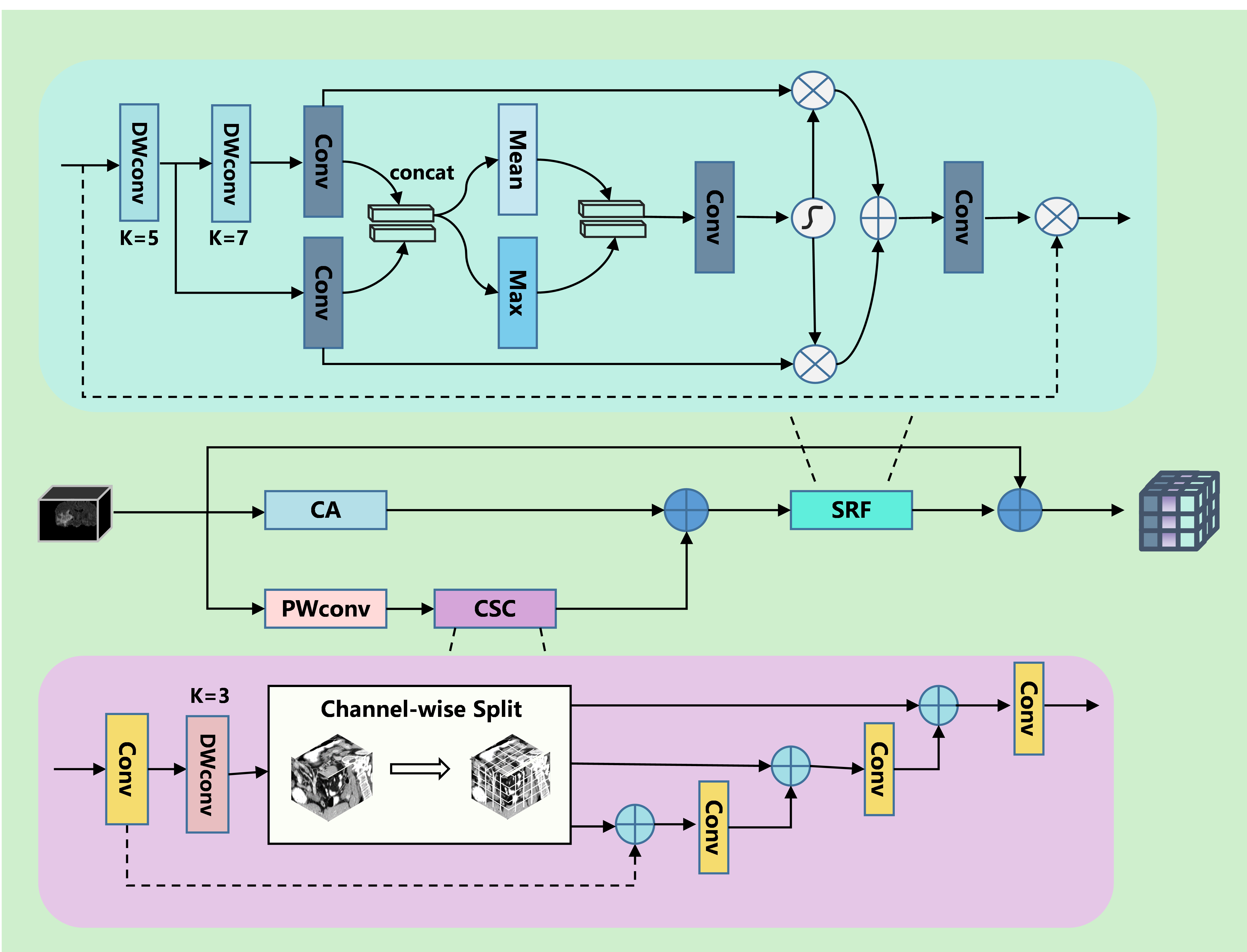}
\caption{Hierarchical Semantic Alignment (HSA) Block integrates three branches: point-wise convolution with Channel Separation Convolution (CSC), Channel Attention (CA), and a residual path with Selective Receptive Field (SRF) module for enhanced feature fusion and refinement.}
\label{fig:HSA}
\end{figure}

\subsection{HSA: Hierarchical Semantic Alignment}
To address the challenges of excessive model parameters and insufficient channel-level attention to tumor-relevant information in 3D segmentation tasks, we propose a HSA block, as shown in Fig. \ref{fig:HSA}. The module enhances semantic consistency and adaptively emphasizes tumor-related modalities or feature maps, thereby achieving lightweight modeling and improving computational efficiency.

The input volume is first subjected to rapid compression and linear transformation to provide more compact semantic features for the subsequent attention module. Important channels or spatial regions—such as tumor boundaries and lesion centers—are then further enhanced. Initial feature transformation yields compressed representations through:

\begin{equation}
\begin{aligned}
\mathcal{\zeta} &= \rho\Big(\mathcal{B}(\mathcal{W}{pw}(\mathcal{F}_{in}))\Big),
\end{aligned}
\end{equation}

\noindent
where $\mathcal{F}_{in}$ denotes the input feature volume, $\mathcal{W}{pw}$ represents a point-wise convolutional operator, $\mathcal{B}(\cdot)$ is batch normalization, and $\rho(\cdot)$ denotes the ReLU activation function. Multi-scale feature fusion is subsequently performed via:

% ===== 定义部分 =====
\begin{equation}
b = \mathcal{S}_b\!\big(\mathcal{P}(\zeta)\big), 
\ \ 
\{e_i\}_{i=0}^{2} = 
\mathrm{Split}\!\left(\mathcal{D}\big(\mathcal{S}_i(\mathcal{P}(\zeta))\big)\right),
\end{equation}

% ===== 递归折叠公式 =====
\begin{equation}
\mathcal{A}^{(1)} \;=\; 
\mathcal{M}_{\mathrm{out}}
\Bigg(
\mathcal{M}_2\!\left(\mathcal{M}_1\!\left(b \oplus e_0 \right) \oplus e_1 \right) \oplus e_2
\Bigg),
\end{equation}

Herein, The projection layer $\mathcal{P}({x})$ expands the input, allowing for complex feature capture. The channel splitting operators $\mathcal{S}_b$ and $\mathcal{S}_i$ extract `base' and `increase' channels, with the latter providing essential fine-grained details. Depthwise separable convolutions $\mathcal{D}$ preserve spatial features efficiently, while the $\mathrm{Split}(\cdot)$ operator divides the output into parts $\{e_0, e_1, e_2\}$. Element-wise addition $\oplus$ and channel mappings $\mathcal{M}_i$ fuse and align multi-scale features, resulting in the final segmentation output $\mathcal{A}^{(1)}$. Channel-wise attention is generated through:

\begin{equation}
\left\{
\begin{aligned}
\mathcal{X}^{(2)} &= \mathcal{P}{\text{avg}}(\mathcal{F}{in}),
\\
\mathcal{B}^{(2)} &= \mathcal{X}^{(2)} \otimes \sigma(\mathcal{C}_{1D}(\mathcal{X}^{(2)})),
\end{aligned}
\right.
\end{equation}

\noindent
where in this branch first performs global spatial context aggregation through average pooling $\mathcal{P}{\text{avg}}(\cdot)$, condensing volumetric information into channel-wise descriptors. A subsequent one-dimensional convolution $\mathcal{C}_{1D}(\cdot)$ models inter-channel dependencies, followed by sigmoid activation $\sigma(\cdot)$ to generate attention weights. Channel-wise reweighting via Hadamard product $\otimes$ adaptively enhances features relevant to hippocampal subfield segmentation, improving sensitivity to intensity variations and structural boundaries in T1-weighted MR volumes. Final feature integration and refinement follows:

\begin{equation}
\left\{
\begin{aligned}
\mathcal{X} &= \mathcal{A}^{(1)} + \mathcal{B}^{(2)},
\\
\mathcal{Z} &= \mathcal{M}_{SRF}(\mathcal{X}; \Psi) + \mathcal{F}{in},
\end{aligned}
\right.
\end{equation}

\noindent
in Eq. (18), the fused feature $\mathcal{X}$ is processed by a multi-scale spatial attention module $\mathcal{M}_{SRF}(\cdot; \Psi)$, which employs depthwise separable convolutions with large kernels (e.g., $5 \times 5 \times 5$, $7 \times 7 \times 7$) to capture anisotropic contextual information across sagittal, coronal, and axial planes. Such a design significantly enlarges the receptive field while maintaining parameter efficiency, enabling precise delineation of hippocampal substructures such as Cornu Ammonis and dentate gyrus despite partial volume effects and low contrast.

\section{EXPERIMENTS}\label{sec:experiments}
This section first introduces the six publicly available medical image datasets used in our study, along with the evaluation metrics employed during the experiments. A comprehensive experimental evaluation of the proposed MSD-KMamba is conducted, followed by analysis and discussion of the results. Furthermore, a visual analysis of the experimental outcomes is provided.

\subsection{Datasets}
To verify the effectiveness of our method, we conducted experiments on the following six datasets, with a particular focus on the segmentation of boundary details within the target regions of all images. The datasets and their quantities are shown in Table \ref{tab:dataset_summary}.

\begin{table}[htbp]
\centering
\caption{Summary of Training, Validation, and Test Cases Across Different Datasets.}
\label{tab:dataset_summary}
\begin{tabular}{cccc}
\toprule
\textbf{Dataset} & \textbf{Train Case} & \textbf{Valid Case} & \textbf{Test Case} \\
\midrule
BraTS 2017 & 228  & 28  & 29  \\
BraTS 2019 & 268  & 32  & 32  \\
GLI 2023   & 500  & 62  & 63  \\
HARP   & 104  & 13  & 13  \\
DecathlonHip   & 208  & 26  & 26  \\
Kulaga-Yoskovitz   & 15  & 4  & 5  \\
\bottomrule
\end{tabular}
\end{table}

% \subsubsection*{1) BraTS 2017, BraTS 2019 
\begin{itemize}
    \item BraTS 2017, BraTS 2019 
\cite{menze2014multimodal},\cite{bakas2017advancing},\cite{bakas2018identifying}:
The Brain Tumor Segmentation (BraTS) 2017 and 2019 challenges provide pre-operative multimodal MRI scans with expert-annotated glioma sub-regions: enhancing tumor (ET), peritumoral edema (ED), and non-enhancing tumor core (NCR/NET). Each case includes four modalities (T1, T1Gd, T2, FLAIR), all skull-stripped, co-registered, and resampled to 1mm³. Evaluation focuses on whole tumor (WT), tumor core (TC), and ET. BraTS 2017 established consistent labeling, while BraTS 2019 expanded data and improved annotations, offering a standardized benchmark for robust algorithm 
development.
    \item GLI 2023 
\cite{baid2021rsna}:
focuses on adult glioma segmentation using mpMRI. ET is identified by hyperintensity on T1Gd, and NCR/NET by low T1Gd and high T2-FLAIR intensity, reflecting tumor heterogeneity. It extends prior versions with a larger multi-institutional dataset and standardized annotations, aiming to improve algorithmic robustness for real-world clinical variability in segmentation and survival prediction.
    \item HARP \cite{boccardi2015training}:
The HarP dataset offers standardized hippocampal segmentations under a harmonized protocol from EADC and ADNI. It includes structural MRI scans across ages and atrophy stages, with Delphi-based manual labels certified as a gold standard. Expert tracers achieved high interrater reliability, ensuring accurate hippocampal volume measurements for algorithm training and validation.
    \item DecathlonHip \cite{simpson2019large}:
Part of the Medical Segmentation Decathlon, DecathlonHip includes 260 T1-weighted MRI scans from healthy and psychotic disorder patients, with expert manual annotations of left/right hippocampi. Data are anonymized, skull-stripped, and normalized. It serves as a standard benchmark for segmenting small brain structures and is widely used in deep learning research.
    \item Kulaga-Yoskovitz \cite{kulaga2015multi}:
This dataset provides high-resolution structural MRI for detailed hippocampal subfield segmentation (e.g., CA1, CA2/3, CA4/DG, subiculum). Its ultra-high resolution enables clear internal architecture visualization and precise 3D expert annotations. It supports fine-grained algorithm validation and probabilistic atlas construction, especially for studies on aging and neurological disorders.
\end{itemize}

\subsection{Evaluation Metrics}
Due to differences in imaging modalities and specific segmentation targets across the six datasets, different evaluation metrics were adopted accordingly. For the BraTS 2017, BraTS 2019, and the GLI 2023, we primarily used the Dice Similarity Coefficient (DSC) as the evaluation metric to assess segmentation performance across three tumor subregions: enhancing tumor (ET), tumor core (TC), and whole tumor (WT).

\begin{equation}
{Dice}(A, B) = \frac{2|A \cap B|}{|A| + |B|},
\end{equation}

\noindent
where $A$ is the set of pixels in the predicted segmentation,
and $B$ is the set of pixels in the ground truth segmentation.

For the HARP dataset, we employed DSC, Hausdorff Distance 95 (HD95), and Intersection over Union (IoU) to comprehensively evaluate the segmentation performance.
For the DecathlonHip dataset and the Kulaga-Yoskovitz hippocampal dataset, we adopted  DSC and HD95 to assess the segmentation quality of different hippocampal subregions.

\begin{equation}
{HD}_{95}(A, B) = \text{quantile}_{95} \left( d(A, B) \right),
\end{equation}

\noindent
where \( d(A, B) \) denotes the set of all shortest distances in pixel units from the predicted boundary \( A \) to the ground truth boundary \( B \), typically computed in both directions to ensure symmetry. The operator \( \text{quantile}_{95} \) refers to the 95th percentile of this distance set, effectively measuring the boundary mismatch while reducing sensitivity to outliers.

\begin{equation}
{IoU}(A, B) = \frac{|A \cap B|}{|A \cup B|},
\end{equation}

\noindent
where \( A \) and \( B \) represent the predicted segmentation and the ground truth segmentation, respectively. \( |A \cap B| \) is the number of overlapping pixels (true positives), and \( |A \cup B| \) is the total number of pixels in both the prediction and ground truth regions. The IoU quantifies the overlap between the predicted and actual segmentation, and is widely used as a performance metric in image segmentation tasks.

\subsection{Implementation Detail}
We implemented our MSD-BMamba model using PyTorch on an RTX A6000 GPU with 48 GB of memory. To ensure consistency across experimental settings, all comparison models and our MSD-BMamba were trained from scratch. For the BraTS 2017, BraTS 2019, and GLI 2023 datasets, the batch size was set to 2, The input tensor is of size $4 \times 128 \times 128 \times 128$, where 4 denotes the number of channels and $128 \times 128 \times 128$ represents the spatial resolution in the 3D volume. During data preprocessing, to mitigate overfitting, we applied random flipping, random cropping, and Gaussian noise augmentation to the input data.

For the Harp, DecathlonHip, and Kulaga-Yoskovitz datasets, the batch size was set to 4, The input volume has a spatial resolution of $64 \times 64 \times 64$ voxels.
 Model training was conducted using the Adam optimizer with a learning rate of 0.001. Throughout the training process, the loss function was a weighted combination of Cross-Entropy Loss and Dice Loss.

\begin{equation}
\mathcal{L}_{origin} = \beta \cdot \mathcal{L}_{CE} + (1 - \beta) \cdot \mathcal{L}_{Dice},
\end{equation}

\noindent
where $\beta$ denotes the trade-off coefficient used to balance the contributions of the Cross-Entropy Loss and the Dice Loss.

When incorporating the self-distillation strategy into our model, the total loss function is defined as the sum of the original loss and the self-distillation loss, expressed as:

\begin{equation}
\mathcal{L}_{total} = \lambda_1 \cdot \mathcal{L}_{origin} + \lambda_2 \cdot \mathcal{L}_{SD},
\end{equation}

\noindent
where $\lambda_1$ is the weighting coefficient for the original loss $\mathcal{L}_{origin}$, and $\lambda_2$ controls the contribution of the self-distillation loss $\mathcal{L}_{SD}$.

\begin{table}[t]
\centering
\caption{Comparison of Dice Scores on the BraTS 2017 Dataset. Bold values indicate the best performance}
\label{tab:brats2017}
\begin{tabular}{ccccc}
\toprule
\textbf{Model} & \textbf{Dice↑} & \textbf{ET↑} & \textbf{TC↑} & \textbf{WT↑} \\
\midrule
TransBTS      & 0.791 & 0.702 & 0.773 & 0.899 \\
UNETR         & 0.785 & 0.700 & 0.763 & 0.892 \\
nnFormer      & 0.774 & 0.664 & 0.756 & 0.901 \\
UNETR++       & 0.811 & 0.724 & 0.801 & 0.910 \\
3D UX-Net      & 0.815 & 0.729 & 0.813 & 0.904 \\
S\textsuperscript{2}CA-Net & 0.809 & 0.714 & 0.809 & 0.905 \\
SegMamba      & 0.812 & 0.719 & 0.814 & 0.903 \\
HCMA-UNet     & 0.781 & 0.683 & 0.768 & 0.894 \\
\textbf{Ours} & \textbf{0.823} & \textbf{0.731} & \textbf{0.823} & \textbf{0.915} \\
\bottomrule
\end{tabular}
\end{table}

\subsection{Comparison With Other Methods}
To validate the performance of MSD-KMamba, we compare it with several state-of-the-art segmentation models, including TransBTS \cite{wang2021transbts}, UNETR \cite{hatamizadeh2022unetr}, nnFormer \cite{zhou2023nnformer}, UNETR++ \cite{shaker2024unetr++}, 3D UX-Net \cite{lee20223d}, S\textsuperscript{2}CA-Net \cite{zhou2024shape}, SegMamba \cite{xing2024segmamba}, and HCMA-UNet \cite{li2025hcma}.

\subsubsection*{1) \textit{Experiments on the BraTS 2017 Segmentation Dataset}}

As shown in Table \ref{tab:brats2017}, the proposed MSD-KMamba demonstrates superior performance across all evaluation metrics on the BraTS 2017 dataset. In particular, it surpasses state-of-the-art baselines such as SegMamba and 3D UX-Net, achieving Dice scores of 0.731, 0.823, and 0.915 for the enhancing tumor (ET), tumor core (TC), and whole tumor (WT), respectively. Moreover, our method achieves the highest average Dice score across all three tumor subregions. These results validate the effectiveness of our model in terms of both segmentation accuracy and boundary delineation, especially in challenging areas like the enhancing tumor.

% \clearpage
\begin{figure*}[!t] % [t] 放在下一页顶部
    \centering
    \includegraphics[width=\textwidth, height=12cm]{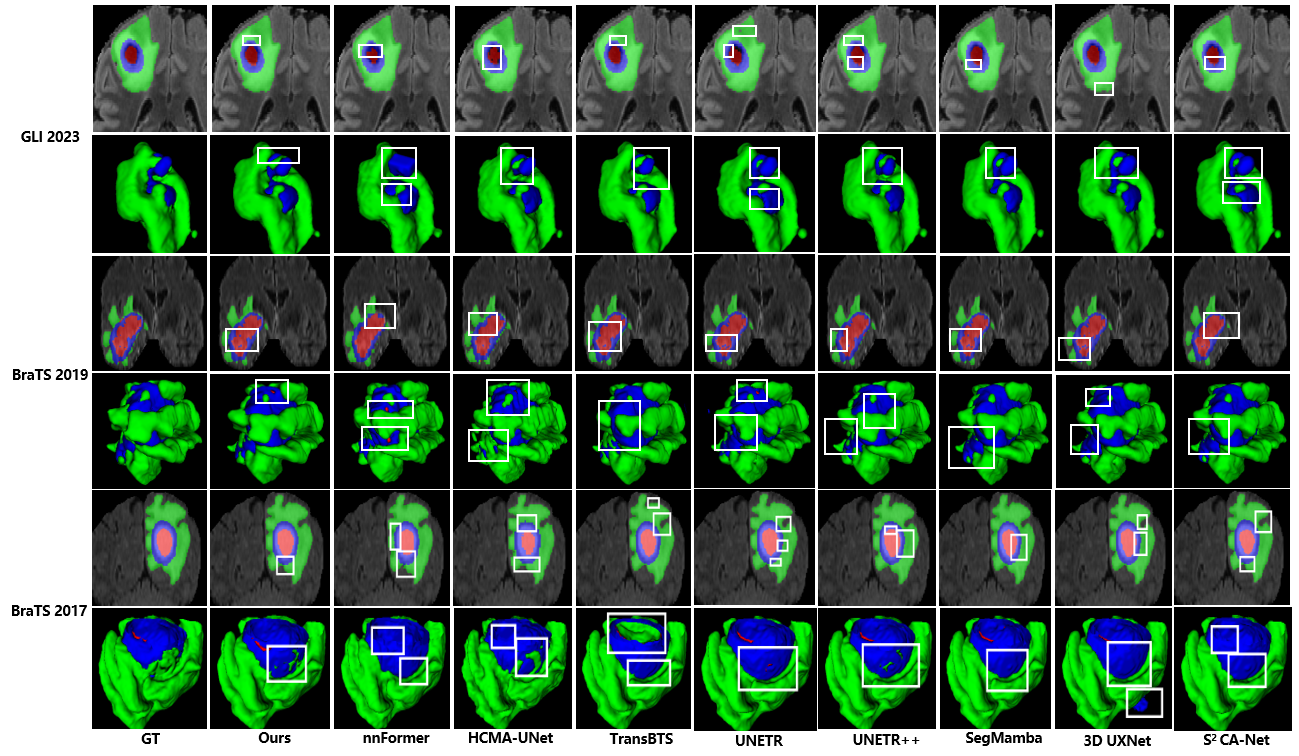}
    \caption{Visual segmentation results of MSD-KMamba and eight comparative methods across three brain tumor datasets, along with corresponding ground truth annotations. Mis-segmented regions are highlighted by white bounding boxes. The dataset is organized in two rows per sample: one row contains 2D slices, and the other contains corresponding 3D patches.}
    \label{fig:slices}
    \vfill % Push caption to bottom
\end{figure*}

In the three brain tumor datasets, red represents the ET region, blue denotes the TC region, and green indicates the WT region. As illustrated in the visual comparison of BraTS 2017 in Fig. \ref{fig:slices}, our method delineates the ET region more clearly, while 3D UX-Net, S\textsuperscript{2}CA-Net, HCMA-UNet, and nnFormer tend to produce relatively blurred boundaries.

\subsubsection*{2) \textit{Experiments on the BraTS 2019 Segmentation Dataset}}
Table \ref{tab:brats2019} presents the segmentation performance on the BraTS 2019 dataset. Our proposed method achieves the highest average Dice score of 0.792, outperforming all comparative methods. The second-best result is achieved by 3D UX-Net with an average Dice score of 0.784. Our approach also demonstrates superior performance in both the ET and WT regions, achieving the best scores in these categories.

\begin{table}[t]
\centering
\caption{Quantitative comparison on the BraTS 2019 dataset.}
\label{tab:brats2019}
\begin{tabular}{ccccc}
\toprule
\textbf{Model} & \textbf{Dice↑} & \textbf{ET↑} & \textbf{TC↑} & \textbf{WT↑} \\
\midrule
TransBTS        & 0.745 & 0.632 & 0.730 & 0.873 \\
UNETR           & 0.734 & 0.640 & 0.708 & 0.854 \\
nnFormer        & 0.754 & 0.692 & 0.723 & 0.848 \\
UNETR++         & 0.774 & 0.698 & 0.749 & 0.878 \\
3D UX-Net        & 0.784 & 0.701 & \textbf{0.779} & 0.874 \\
S\textsuperscript{2} CA-Net & 0.753 & 0.662 & 0.742 & 0.875 \\
SegMamba        & 0.777 & 0.684 & 0.771 & 0.878 \\
HCMA-UNet       & 0.743 & 0.627 & 0.733 & 0.871 \\
\textbf{Ours}   & \textbf{0.792} & \textbf{0.720} & 0.776 & \textbf{0.881} \\
\bottomrule
\end{tabular}
\end{table}

It is worth noting that the Dice scores for the WT region produced by TransBTS, UNETR++, 3D UX-Net, S\textsuperscript{2}CA-Net, SegMamba, and HCMA-UNet are relatively close to each other. This is primarily because the WT region contains the largest number of positive voxels, making the Dice score less sensitive to minor segmentation errors. In contrast, the ET and TC regions contain fewer voxels, and are therefore more sensitive to prediction deviations, resulting in larger performance fluctuations.

As shown in the visual comparison of BraTS 2019 in Fig. \ref{fig:slices}, the WT region exhibits a complex topological structure in 3D space. Accurate segmentation requires capturing spatial dependencies; otherwise, it can lead to discontinuous regions. Our model demonstrates a superior capability in modeling such long-range spatial dependencies compared to other methods.

\subsubsection*{3) \textit{Experiments on the GLI 2023 Segmentation Dataset}}

\begin{table}[t]
\centering
\caption{Comparison of Dice Scores on the GLI 2023 Dataset}
\label{tab:GLI2023}
\begin{tabular}{ccccc}
\toprule
\textbf{Model} & \textbf{Dice↑} & \textbf{ET↑} & \textbf{TC↑} & \textbf{WT↑} \\
\midrule
TransBTS        & 0.902 & 0.877 & 0.908 & 0.923 \\
UNETR           & 0.871 & 0.832 & 0.859 & 0.922 \\
nnFormer        & 0.871 & 0.818 & 0.880 & 0.916 \\
UNETR++         & 0.895 & 0.864 & 0.897 & 0.926 \\
3D UX-Net        & 0.900 & 0.868 & 0.894 & \textbf{0.940} \\
S$^2$ CA-Net    & 0.890 & 0.860 & 0.888 & 0.923 \\
SegMamba        & 0.900 & 0.871 & 0.900 & 0.930 \\
HCMA-UNet       & 0.887 & 0.849 & 0.888 & 0.924 \\
\textbf{Ours}   & \textbf{0.909} & \textbf{0.880} & \textbf{0.913} & 0.935 \\
\bottomrule
\end{tabular}
\end{table}

Results on the GLI 2023 dataset are presented in Table \ref{tab:GLI2023}. Our model achieves the best performance in terms of average Dice score, as well as on the ET and TC regions. It ranks first in three out of the four evaluated metrics, particularly excelling in the segmentation of the ET and TC regions, which are considered the most challenging due to their small volume and complex structure. These results demonstrate the superior capability of our method in capturing fine-grained details and complex spatial dependencies.

As illustrated in Fig. \ref{fig:slices}, Compared to other models, the proposed approach yields more accurate delineation of tumor subregions, particularly the ET and TC areas, which are typically more challenging due to their small volume and irregular morphology. The 3D visualizations further highlight that our model preserves spatial continuity and avoids artifacts, as indicated by the highlighted boxes.

% \clearpage
\begin{figure*}[!t] % [t] 放在下一页顶部
    \centering
    \includegraphics[width=\textwidth, height=12cm]{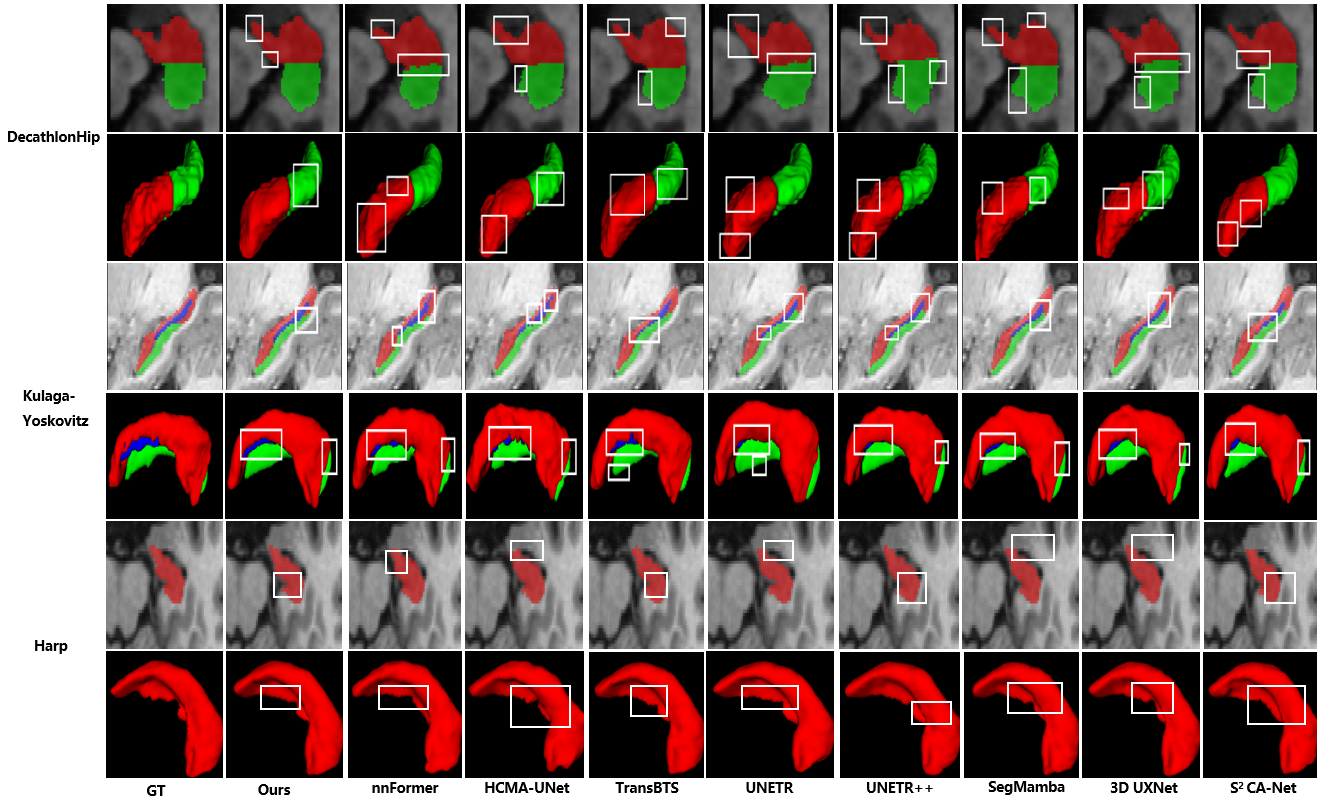}
    \caption{Visual segmentation results of MSD-KMamba and eight comparative methods across three hippocampal datasets, along with corresponding ground truth annotations. Mis-segmented regions are highlighted by white bounding boxes. The dataset is organized in two rows per sample: one row contains 2D slices, and the other contains corresponding 3D patches.}
    \label{fig:slices2}
    \vfill % Push caption to bottom
\end{figure*}

\begin{table}[t]
\centering
\caption{Comparison of 3D HARP Hippocampus Segmentation Performance on Dice, HD95, and IoU}
\label{tab:harphippocampus}
\renewcommand{\arraystretch}{1.2}
\setlength{\tabcolsep}{5pt}
\begin{tabular}{cccc}
\toprule
\textbf{Model} & \textbf{Dice $\uparrow$} & \textbf{HD95 $\downarrow$} & \textbf{IoU $\uparrow$} \\
\midrule
TransBTS       & 0.8785 & 1.7444 & 0.7841 \\
UNETR          & 0.8674 & 1.3603 & 0.7674 \\
nnFormer       & 0.8439 & 2.0546 & 0.7312 \\
UNETR++        & 0.8778 & 1.0882 & 0.7829 \\
3D UX-Net       & 0.8905 & 1.0637 & 0.8032 \\
S$^2$ CA-Net   & 0.8707 & 1.0882 & 0.7715 \\
SegMamba       & 0.8975 & \textbf{1.0000} & 0.8143 \\
HCMA-UNet      & 0.8597 & 1.1912 & 0.7544 \\
\textbf{Ours}  & \textbf{0.9025} & \textbf{1.0000} & \textbf{0.8226} \\
\bottomrule
\end{tabular}
\end{table}

\subsubsection*{4) \textit{Experiments on the HARP Segmentation Dataset}}
As shown in Table \ref{tab:harphippocampus}, our proposed method achieves the best 

\noindent
performance across all three metrics on the 3D hippocampus segmentation task. Specifically, it obtains the highest Dice score (0.9025) and IoU (0.8226), and ties with SegMamba for the lowest HD95 (1.0000). These results indicate the strong capability of our method in accurately capturing both the overall structure and fine boundary details of the hippocampus, especially in handling its complex and elongated morphology.

As shown in Fig. \ref{fig:slices2}, the proposed method achieves more accurate and consistent hippocampus segmentation compared to eight state-of-the-art baselines. In both 2D axial slices and 3D renderings, our results show better alignment with the ground truth, particularly in challenging boundary regions, as highlighted by the white boxes. This demonstrates the robustness and structural fidelity of our approach in capturing fine-grained anatomical details.

\begin{table}[!t]
\centering
\caption{Comparison of segmentation performance in Decathlon anterior and posterior hippocampus regions.}
\begin{tabular}{cccccc}
\toprule
\multirow{2}{*}{\textbf{Model}} & \multicolumn{2}{c}{\textbf{Anterior}} & \multicolumn{2}{c}{\textbf{Posterior}} \\
\cmidrule(lr){2-3} \cmidrule(lr){4-5}
 & Dice$\uparrow$ & HD95$\downarrow$ & Dice$\uparrow$ & HD95$\downarrow$ \\
\midrule
TransBTS      & 0.8750 & 1.2717 & 0.8391 & 1.6302 \\
UNETR         & 0.8719 & 1.3785 & 0.8398 & 1.6102 \\
nnFormer      & 0.8582 & 1.3170 & 0.8298 & 1.4162 \\
UNETR++       & 0.8698 & 1.2457 & 0.8434 & 1.3351 \\
3D UX-Net      & 0.8718 & 1.3555 & 0.8438 & 1.5218 \\
S$^{2}$ CA-Net& 0.8733 & 1.2494 & 0.8425 & 1.5357 \\
SegMamba      & 0.8830 & 1.2016 & 0.8522 & 1.6743 \\
HCMA-UNet     & 0.8702 & 1.1847 & 0.8322 & 1.4293 \\
\textbf{Ours} & \textbf{0.8868} & \textbf{1.1022} & \textbf{0.8605} & \textbf{1.2906} \\
\bottomrule
\end{tabular}
\label{tab:DecathlonHip}
\end{table}

\subsubsection*{5) \textit{Experiments on the DecathlonHip Segmentation Dataset}}
As shown in Table \ref{tab:DecathlonHip}, the proposed method achieves the best performance in both anterior and posterior regions of the hippocampus on the DecathlonHip dataset. Specifically, we obtain the highest Dice scores of 0.8868 and 0.8605, and the lowest HD95 distances of 1.1022 and 1.2906 in the anterior and posterior regions, respectively. 

3D UX-Net achieves a posterior Dice score of 0.8438; however, the HD95 reaches 1.5218, indicating instability in boundary delineation. SegMamba attains a high anterior Dice score of 0.8830 (close to the best), but its posterior HD95 is as high as 1.6743, suggesting poor structural continuity.

As shown in Fig. \ref{fig:slices2}, the red and green regions represent the anterior and posterior hippocampus, respectively. Our method demonstrates superior performance in terms of boundary preservation, spatial consistency, and structural integrity. Notably, it achieves a more accurate reconstruction of the posterior hippocampus compared to the ground truth (GT), effectively avoiding common issues observed in other methods, such as fragmentation, misalignment, and over-smoothing.

\subsubsection*{6) \textit{Experiments on the Kulaga-Yoskovitz Segmentation Dataset}}
As shown in Table \ref{tab:Kulaga-Yoskovitz}, the overall performance of the UNETR series is suboptimal, primarily due to insufficient modeling of spatial structures. While SegMamba and 3D UX-Net achieve competitive Dice scores, their relatively high HD95 values indicate deficiencies in structural continuity and boundary smoothness. In contrast, the proposed method consistently outperforms existing state-of-the-art approaches across all subfields, demonstrating superior performance in structural consistency, boundary accuracy, and fine-grained anatomical reconstruction.

\begin{table*}[!t]
\centering
\caption{Comparison of segmentation performance on Kulaga-Yoskovitz hippocampus dataset.}
% 适当增加列间距和行距
\setlength{\tabcolsep}{12pt} % 默认一般 6pt，可以改成 8~10pt
\renewcommand{\arraystretch}{1.2} % 行高稍微加大

\begin{tabular}{ccccccc}
\toprule
\multirow{2}{*}{\textbf{Model}} & \multicolumn{2}{c}{\textbf{CA1-3}} & \multicolumn{2}{c}{\textbf{Sub}} & \multicolumn{2}{c}{\textbf{CA4-DG}} \\
\cmidrule(lr){2-3} \cmidrule(lr){4-5} \cmidrule(lr){6-7}
 & Dice$\uparrow$ & HD95$\downarrow$ & Dice$\uparrow$ & HD95$\downarrow$ & Dice$\uparrow$ & HD95$\downarrow$ \\
\midrule
TransBTS      & 0.7910 & 1.4485 & 0.7404 & 1.4485 & 0.7464 & 1.9128 \\
UNETR         & 0.7024 & 2.2714 & 0.5726 & 2.4472 & 0.6564 & 2.4490 \\
nnFormer      & 0.7608 & 1.6585 & 0.6900 & 1.9237 & 0.7114 & 1.4485 \\
UNETR++       & 0.7855 & 1.3314 & 0.7482 & 1.3314 & 0.7526 & 1.8601 \\
3D UX-Net      & 0.8159 & 1.2485 & 0.7519 & 1.4485 & 0.7762 & 1.5384 \\
S$^{2}$ CA-Net& 0.7963 & 1.3121 & 0.7475 & 1.3121 & 0.7572 & 1.6585 \\
SegMamba      & 0.8241 & 1.1657 & 0.7672 & 1.3657 & 0.7969 & 1.3301 \\
HCMA-UNet     & 0.7478 & 1.7121 & 0.6762 & 1.7657 & 0.7123 & 1.9852 \\
\textbf{Ours} & \textbf{0.8499} & \textbf{1.0000} & \textbf{0.8055} & \textbf{1.0000} & \textbf{0.8185} & \textbf{1.1657} \\
\bottomrule
\end{tabular}
\label{tab:Kulaga-Yoskovitz}
\end{table*}

As shown in Fig. \ref{fig:slices2}, which visualizes the segmentation results on the Kulaga-Yoskovitz dataset, the red, blue, and green regions represent CA1-3, CA4-DG, and Subiculum of a single hippocampus, respectively. Both the ground truth and the proposed method exhibit a continuous and complete hippocampal structure, particularly maintaining anatomical integrity in the tail and curved regions. In contrast, other models, such as UNETR and nnFormer, suffer from boundary discontinuities or local structural distortions, especially in the red  and green regions. 

\begin{table}[!t]
\centering
\caption{ABLATION EXPERIMENTS ON THE DecathlonHip dataset.}
\begin{tabular}{ccccc}
\toprule 
\multirow{2}{*}{\textbf{Model}} & \multicolumn{2}{c}{\textbf{Anterior}} & \multicolumn{2}{c}{\textbf{Posterior}} \\
\cmidrule(lr){2-3} \cmidrule(lr){4-5}
 & Dice$\uparrow$ & HD95$\downarrow$ & Dice$\uparrow$ & HD95$\downarrow$ \\
\midrule
Baseline             & 0.8805 & 1.2332 & 0.8496 & 1.6290 \\
+HSA                 & 0.8824 & 1.1463 & 0.8512 & 1.4745 \\
+BKM                 & 0.8820 & 1.1845 & 0.8512 & 1.6454 \\
+MDA                 & 0.8806 & 1.2203 & 0.8523 & 1.6221 \\
+HSA+BKM             & 0.8842 & 1.1247 & 0.8539 & 1.5031 \\
+HSA+MDA             & 0.8825 & 1.2139 & 0.8589 & 1.3502 \\
+BKM+MDA             & 0.8856 & 1.1538 & 0.8580 & 1.4555 \\
\textbf{+HSA+BKM+MDA} & \textbf{0.8868} & \textbf{1.1022} & \textbf{0.8605} & \textbf{1.2906} \\
\bottomrule
\end{tabular}
\label{tab:ablation-decathlonhip}
\end{table}

\subsection{Ablation Studies}
Our approach primarily incorporates three modules: Hierarchical Semantic Alignment
(HSA) module, the Bidirectional Kan-enhanced Mamba (BKM) module, and the Multi-scale Self-distillation Aggregation (MDA) module. Ablation experiments were performed on the DecathlonHip datasets.

As shown in Tables \ref{tab:ablation-decathlonhip}, the combination “Baseline + HSA + BKM + MDA” consistently outperforms all other variants across all datasets, demonstrating the effectiveness of the proposed modular design.
Specifically, when adding any one of the HSA, BKM, or MDA modules individually, notable performance gains are observed in both the anterior and posterior hippocampus segmentation results on the DecathlonHip datasets. Further improvements are achieved by combining the modules in pairs. Finally, the integration of all three modules—HSA, BKM, and MDA—yields the best overall segmentation performance.

These results also emphasize the critical role of enhancing feature extraction capability within the encoder–decoder framework.

% \begin{table}[!t]
% \centering
% \caption{ABLATION EXPERIMENTS ON THE DecathlonHip dataset.}
% \begin{tabular}{ccccc}
% \toprule 
% \multirow{2}{*}{\textbf{Model}} & \multicolumn{2}{c}{\textbf{Anterior}} & \multicolumn{2}{c}{\textbf{Posterior}} \\
% \cmidrule(lr){2-3} \cmidrule(lr){4-5}
%  & Dice$\uparrow$ & HD95$\downarrow$ & Dice$\uparrow$ & HD95$\downarrow$ \\
% \midrule
% Baseline             & 0.8805 & 1.2332 & 0.8496 & 1.6290 \\
% +HSA                 & 0.8824 & 1.1463 & 0.8512 & 1.4745 \\
% +BKM                 & 0.8820 & 1.1845 & 0.8512 & 1.6454 \\
% +MDA                 & 0.8806 & 1.2203 & 0.8523 & 1.6221 \\
% +HSA+BKM             & 0.8842 & 1.1247 & 0.8539 & 1.5031 \\
% +HSA+MDA             & 0.8825 & 1.2139 & 0.8589 & 1.3502 \\
% +BKM+MDA             & 0.8856 & 1.1538 & 0.8580 & 1.4555 \\
% \textbf{+HSA+BKM+MDA} & \textbf{0.8868} & \textbf{1.1022} & \textbf{0.8605} & \textbf{1.2906} \\
% \bottomrule
% \end{tabular}
% \label{tab:ablation-decathlonhip}
% \end{table}

\begin{figure}[!t]
    \centering
    \includegraphics[width=0.9\linewidth,height=0.33\textheight]{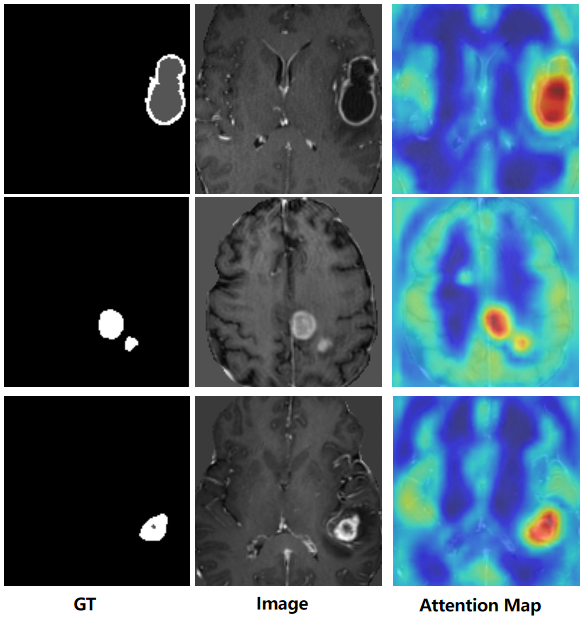}
    \caption{The visualization results of attention on GLI 2023 datasets.}
    \label{fig:attentionmap}
\end{figure}

\subsection{Attention Mechanism Visualization}
In our proposed model, we incorporate an attention-based mechanism within the HSA module. To intuitively demonstrate the impact of these modules in segmenting specific tumor regions, we perform attention heatmap visualization, as illustrated in Fig. \ref{fig:attentionmap}. The GLI 2023 dataset, which contains four MRI modalities (T1, T1Gd, T2, and FLAIR), is employed for this analysis. The visualization highlights the attention responses learned by the model, providing insight into how the attention modules contribute to the segmentation process.

In the first row, the attention map highlights a strong response in red over the tumor region, indicating that the attention mechanism effectively captures highly relevant features in this area. In the second row, the attention region fully covers both tumor lesions, while the surrounding blue-green background indicates weak responses in non-lesion areas, demonstrating the model’s strong discriminative capability. In the third row, despite the small size of the lesion, the attention module successfully identifies and localizes it, showing the model’s sensitivity to fine-grained features. These results suggest that the attention mechanism significantly enhances the model’s ability to discriminate and localize target regions in medical image segmentation tasks, particularly under complex backgrounds and multi-modal inputs.

% \clearpage

\begin{figure}[t]
    \centering
    \includegraphics[width=1\linewidth,height=0.29\textheight]{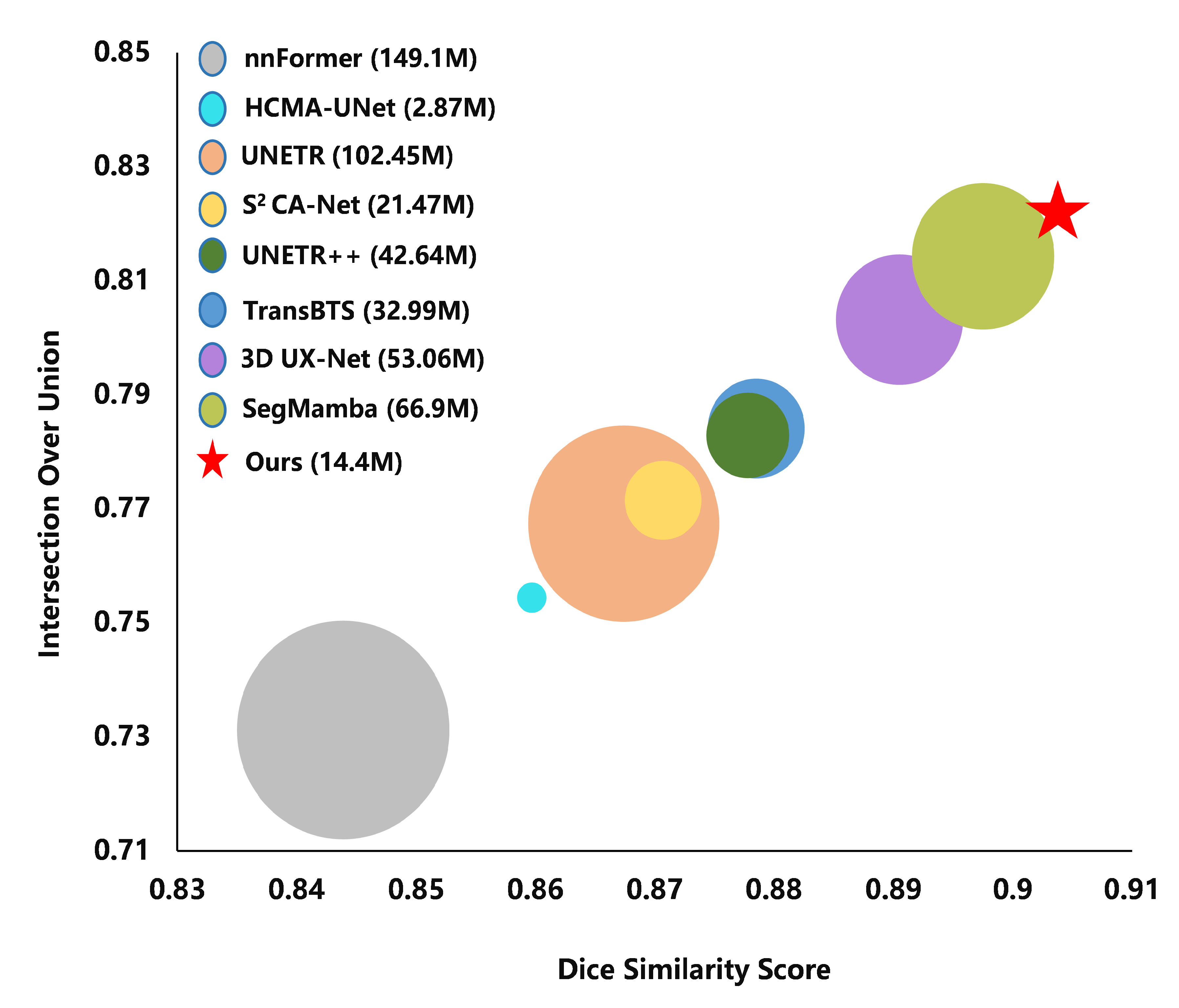}
    \caption{Accuracy vs. model complexity comparison on the HARP dataset.}
    \label{fig:bubble}
\end{figure}
\subsection{Discussion}
To further validate these experimental results and explore their potential value in real clinical applications, we conduct the following series of discussions:

\subsubsection*{1) \textit{Discussion on the Model Complexity}}

In our study, we emphasize the trade-off among model performance, architectural complexity. To this end, we conduct a comprehensive set of experiments in which our proposed model is thoroughly compared against eight state-of-the-art counterparts.

As illustrated in Fig. \ref{fig:bubble}, the proposed model achieves the highest Dice similarity coefficient and Intersection over Union on the Harp hippocampus segmentation dataset, while maintaining a compact architecture with only 12.4 million parameters. This performance significantly outperforms larger models such as SegMamba (66.9M), UNETR (120.6M), and nnFormer (149.1M), demonstrating an excellent balance between segmentation accuracy and computational efficiency.

In comparison, SegMamba also yields competitive performance, but requires more than four times the parameters. Models such as UX-Net (53.06M) and TransBTS (31.1M) achieve moderate performance but incur higher or comparable complexity. Notably, although UNETR and nnFormer contain a large number of parameters, their segmentation performance remains inferior, indicating that increased model size does not necessarily guarantee improved results. Meanwhile, HCMA-UNet, a highly compact model with only 2.87M parameters, suffers from limited segmentation accuracy, making it suitable for lightweight applications but less ideal in clinical settings that demand high precision.

Overall, the proposed method effectively optimizes the model architecture to achieve a well-balanced trade-off among accuracy, model complexity, highlighting its strong potential for deployment in real-world clinical practice.

% \begin{figure}[t]
%     \centering
%     \includegraphics[width=1\linewidth,height=0.3\textheight]{fig/bubble.png}
%     \caption{Accuracy vs. model complexity comparison on the HARP dataset.}
%     \label{fig:bubble}
% \end{figure}

\subsubsection*{2) \textit{Discussion on the self-distillation loss}}

\begin{figure}[t]
    \centering
    \includegraphics[width=1\linewidth,height=0.28\textheight]{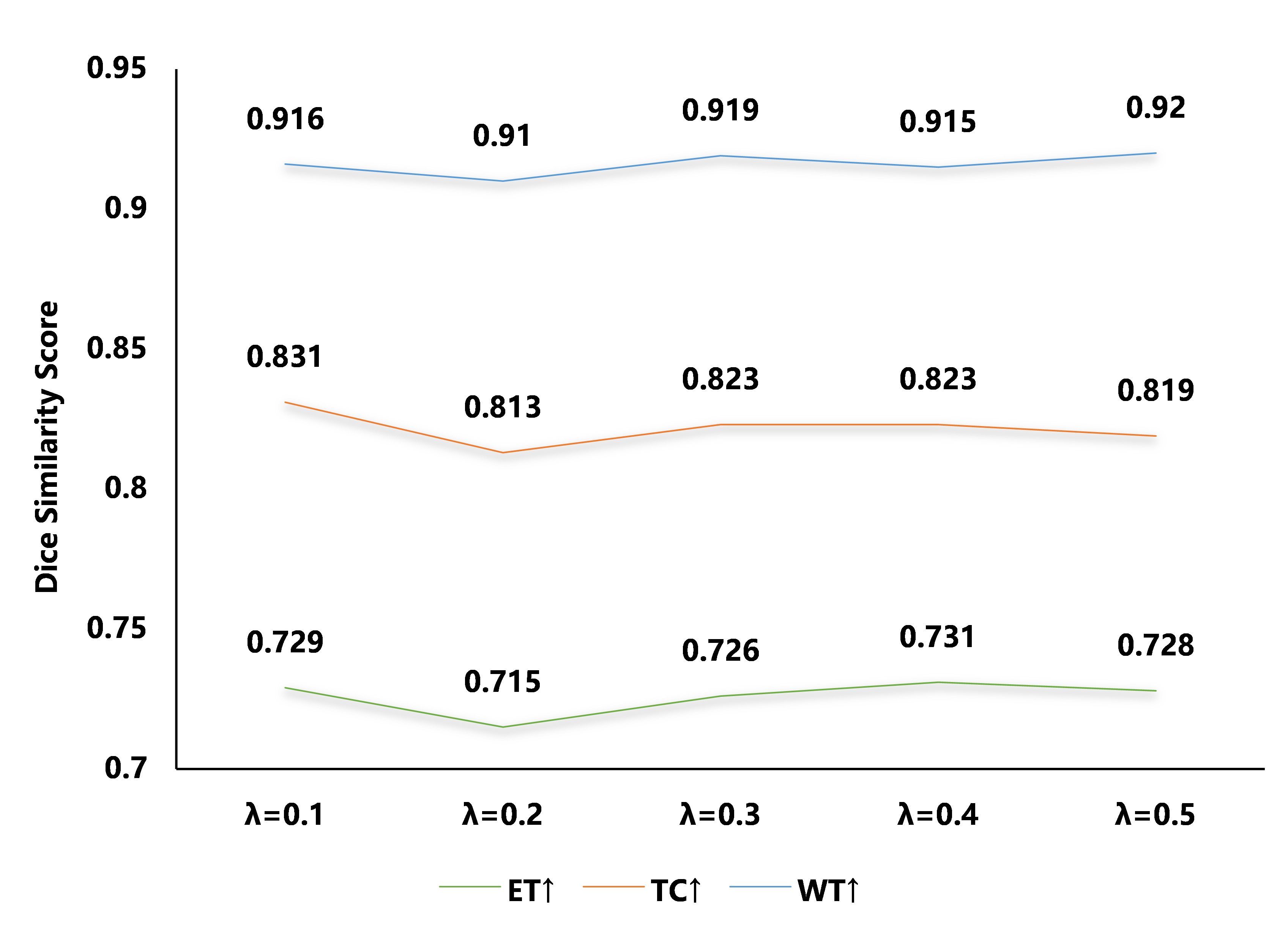}
    \caption{Performance comparison with varying distillation loss weight $\lambda$ on the BraTS 2017 dataset. The segmentation accuracy is evaluated using Dice coefficients across three tumor subregions: Enhancing Tumor (ET), Tumor Core (TC), and Whole Tumor (WT).}
    \label{fig:dist}
\end{figure}

To investigate the influence of the distillation loss weighting on segmentation performance, we evaluate various values of the coefficient $\lambda$ on the BraTS 2017 dataset, As shown in Fig. \ref{fig:dist}, $\lambda = 0.1$ yields the best overall performance, achieving the highest Dice score for the TC region (0.831) and competitive results in the ET and WT regions. This demonstrates a well-balanced segmentation capability across different tumor subregions, making it a suitable default configuration for general clinical use. In contrast, $\lambda = 0.2$ exhibits the weakest performance across all three regions, indicating that an insufficient distillation signal may hinder the student model’s ability to effectively learn from the teacher model.

Meanwhile, $\lambda = 0.3$ and $\lambda = 0.4$ demonstrate relatively stable and balanced performance, with $\lambda = 0.4$ achieving the best result for the ET region (0.731), suggesting its applicability in scenarios prioritizing accurate detection of enhancing tumor areas. On the other hand, $\lambda = 0.5$ yields the highest Dice score for the WT region (0.920), making it more suitable for applications requiring comprehensive tumor region coverage. These results suggest that the choice of $\lambda$ should be tailored to the specific segmentation objective, balancing precision in critical regions against overall robustness.

\begin{figure}[t]
    \centering
    \includegraphics[width=1\linewidth,height=0.3\textheight]{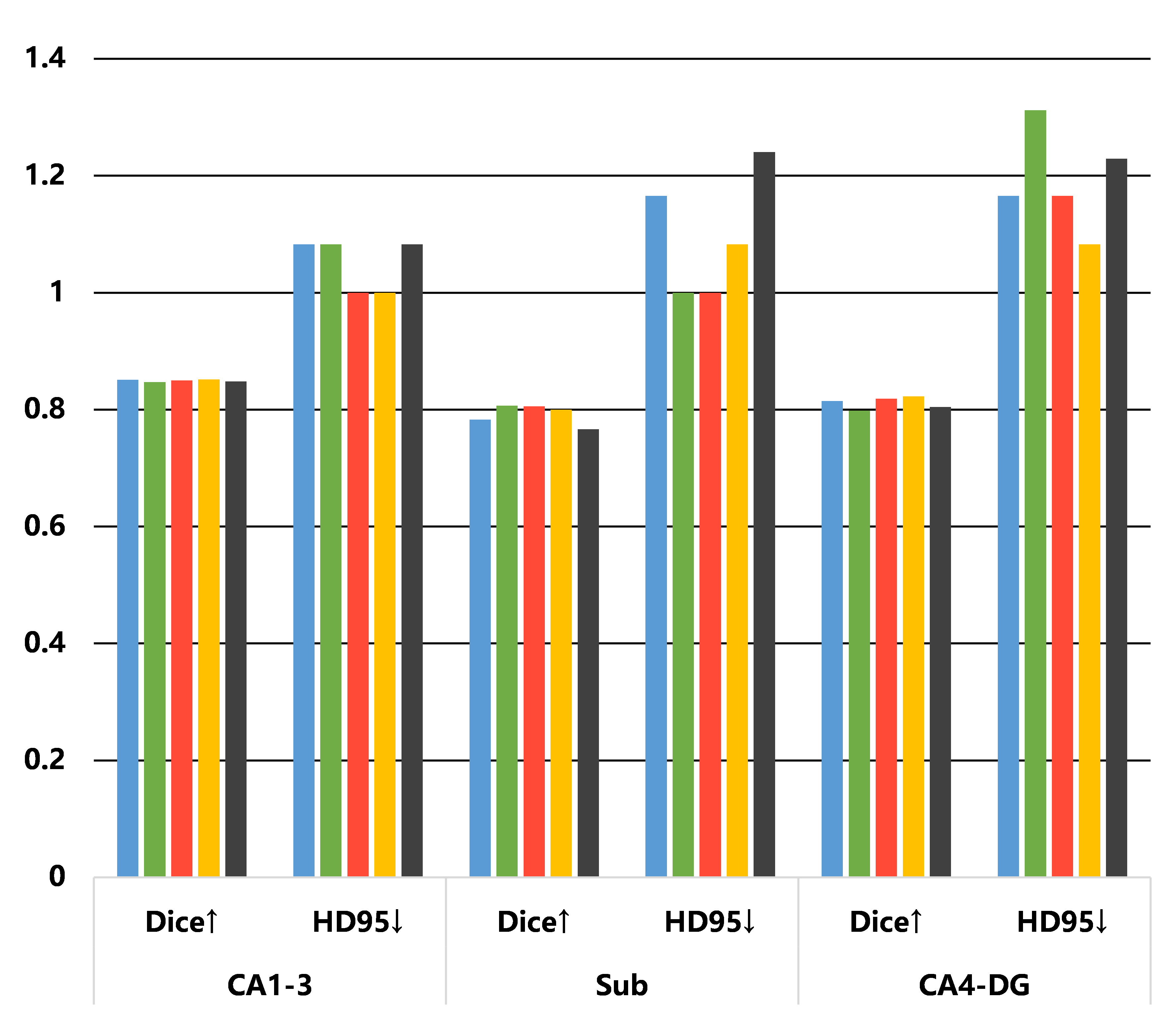}
    \caption{Effect of hidden layer width on segmentation performance in the KAN network evaluated on the Kulaga-Yoskovitz dataset. Results are reported as Dice (↑) and HD95 (↓) for CA1-3, Subiculum, and CA4-DG. The parameter L denotes the width of the hidden layer in the KAN network.}
    \label{fig:kanhidden}
\end{figure}

\subsubsection*{3) \textit{Discussion on the impact of hidden layer within the KAN network}}
To investigate the impact of hidden layer width on segmentation performance within the KAN network, we evaluate five different layer sizes ($L = 8, 16, 32, 64, 128$) on the Kulaga-Yoskovitz dataset. As shown in Fig.~\ref{fig:kanhidden}, when $L=64$, the model achieves the best overall balance across all three hippocampal subregions (CA1-3, Subiculum, and CA4-DG). Notably, the Dice scores for CA1-3 (0.8515) and CA4-DG (0.8228) are the highest, and the HD95 values remain low, indicating accurate boundary delineation. 

Although some configurations like $L=16$ and $L=32$ also yield competitive results in certain subregions, their performance is less consistent. In contrast, setting $L=128$ leads to a noticeable drop in performance for the Subiculum (Dice: 0.7662, 
 HD95: 1.2402), suggesting possible overfitting or increased model complexity beyond optimal capacity. These results highlight that an intermediate hidden layer width, specifically $L=64$, provides a favorable trade-off between representation capability and generalization performance in hippocampal subfield segmentation.

\section{CONCLUSION}\label{sec:conclusion}
In this study, a novel three-dimensional brain region segmentation framework is proposed, integrating a multi-scale contextual self-distillation mechanism with a hybrid backbone architecture combining Kolmogorov–Arnold representations and state space models. The self-distillation strategy facilitates the transfer of both semantic and structural knowledge from the teacher network to the student network, thereby enhancing model generalization and segmentation efficiency. This capability is particularly vital when dealing with complex and heterogeneous 3D medical imaging data, such as brain tumors and hippocampal structures. Furthermore, the hybrid backbone enhances the model’s representational power: the Kolmogorov–Arnold module delivers efficient nonlinear mapping with high parameter efficiency, while the state space model excels at capturing long-range contextual dependencies with low computational overhead. This combined architecture effectively integrates local texture features and global semantic information. Extensive evaluations on six public benchmarks confirm the method’s superior performance across multiple metrics, highlighting its robustness and clinical applicability for precise segmentation of brain regions.

Despite the method’s competitive performance and lightweight design, the self-distillation component exhibits sensitivity to hyperparameter configurations, requiring careful calibration per dataset to achieve stable results. Additionally, the approach depends on soft labels generated by the teacher model. Inaccurate predictions from the teacher may propagate errors to the student network, potentially degrading segmentation quality. To overcome these limitations, future work will investigate dynamic weighting mechanisms that adaptively balance the influence of self-distillation based on training progression and uncertainty estimation. Further research will also aim to develop safeguards against misleading supervisory signals from the teacher, with the goal of improving diagnostic accuracy and therapeutic planning. It is anticipated that this framework will contribute significantly to the development of intelligent medical imaging systems.

% 参考文献
\bibliography{references}
% \bibliographystyle{IEEEtran}   % 指定样式
% \bibliography{references.bib}           % mybib.bib 文件里放条目
%                                  % 这行会自动生成“REFERENCES”小节标题

\end{document}